# Detection of Adulteration in Coconut Milk using Infrared Spectroscopy and Machine Learning

Mokhtar A. Al-Awadhi  
*Department of Computer Science and IT*  
*Dr. Babasaheb Ambedkar Marathwada University*  
Aurangabad, India  
mokhtar.awadhi@gmail.com

Ratnadeep R. Deshmukh  
*Department of Computer Science and IT*  
*Dr. Babasaheb Ambedkar Marathwada University*  
Aurangabad, India  
rrdeshmukh.csit@bamu.ac.in

*Abstract*— In this paper, we propose a system for detecting adulteration in coconut milk, utilizing infrared spectroscopy. The machine learning-based proposed system comprises three phases: preprocessing, feature extraction, and classification. The first phase involves removing irrelevant data from coconut milk spectral signals. In the second phase, we employ the Linear Discriminant Analysis (LDA) algorithm for extracting the most discriminating features. In the third phase, we use the K-Nearest Neighbor (KNN) model to classify coconut milk samples into authentic or adulterated. We evaluate the performance of the proposed system using a public dataset comprising Fourier Transform Infrared (FTIR) spectral information of pure and contaminated coconut milk samples. Findings show that the proposed method successfully detects adulteration with a cross-validation accuracy of 93.33%.

*Keywords—Coconut Milk Adulteration, FTIR Spectroscopy, Linear Discriminant Analysis, K-Nearest Neighbors, Machine Learning*

## I. INTRODUCTION

Financially motivated adulteration of liquid food has recently become a global problem. This deceptive and deliberate food tampering has resulted in hundreds of fatalities and disease outbreaks [1]. Because of a consumer's inability to recognize the authenticity of a product through visual inspection or smell, the contaminated product frequently goes undetected.

Coconut milk, which is one of the liquid food vulnerable to adulteration, is a milky oil-in-water emulsion made from coconut flesh aqueous extract [2]. Coconut milk is considerably better than other saturated fat products because it is rich in saturated fat, particularly as medium-chain fatty acids, which the body can easily absorb [3]. However, tampering with coconut milk can lead to health problems that raise the risk of death and morbidity. Therefore, it is vital to analyze coconut milk for adulterants that lower its quality.

Chemical-based methods can detect adulteration in coconut milk; however, these methods are destructive and time-consuming. Modern techniques for detecting adulteration in coconut milk are fast and noninvasive. These techniques include infrared (IR) spectroscopy and machine learning (ML). IR spectroscopy is a technique that measures absorbance of IR radiation by molecules in a mixture [4]. The absorbance spectrum can provide a fingerprint for characterizing mixture components. Few studies used IR spectroscopy with ML techniques to detect adulteration in coconut milk. For instance, researchers in [5] applied a partial least squares regression (PLSR) model on FTIR coconut milk spectral data to detect adulteration with corn flour. The created model suited the data well. It achieved a determination coefficient ($R^2$) of 0.9982 and a root-mean-squared calibration error (RMSEC) of 0.688.

The objectives of the present paper are to develop an ML-based method for fast detection of coconut milk adulteration with water. The developed method provides automatic and nondestructive detection of adulteration using infrared spectroscopic data. The dataset used in this research comprises FTIR spectral data of 42 coconut milk samples. The specimens include 14 pure, 14 adulterated with water at 10% concentration, and 14 mixed with water at 20% concentration [6]. The coconut milk samples came from two different sources: the first type was purchased from traditional markets, while another type was instant milk purchased from modern markets. The data creator used an FTIR spectrometer to acquire the spectral data of the coconut milk samples. Each dataset instance comprises 729 features, which represent spectral bands ranging from 2500 to 4000 nm. There are three class labels in the dataset, which are authentic, adulterated10, and adulterated20. Figure 1 shows the infrared spectra of some instances from each class in the dataset. The figure shows the absorbance values of coconut milk in different bands. We can observe that the absorbance increases when the amount of added water increases.

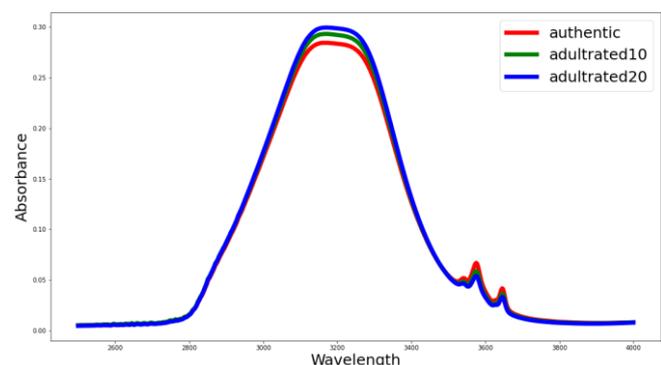

Fig. 1. Spectral data of some instances from different classes in the dataset

The previous work on this dataset used two regression models, namely PCR and PLSR, to detect and quantify the adulteration [6]. The models applied to the two coconut milk types achieved a coefficient of determination $R^2$ of 0.97 for the coconut milk samples from the traditional markets. For the instant coconut milk samples, the PCR and PLSR models got a coefficient of determination $R^2$ of 0.89 and

0.93, respectively. Unlike previous work, which used two prediction models for each milk type separately, we build, in this study, one ML model capable of predicting adulteration in both coconut milk types. We apply supervised ML classification algorithms, such as KNN and support vector machines (SVM), to discriminate between authentic and adulterated coconut milk. The models also predict the two adulteration levels.

## II. PROPOSED SYSTEM

The system proposed in the present research comprises three main phases, as shown in figure 2, which are preprocessing, feature extraction, and classification.

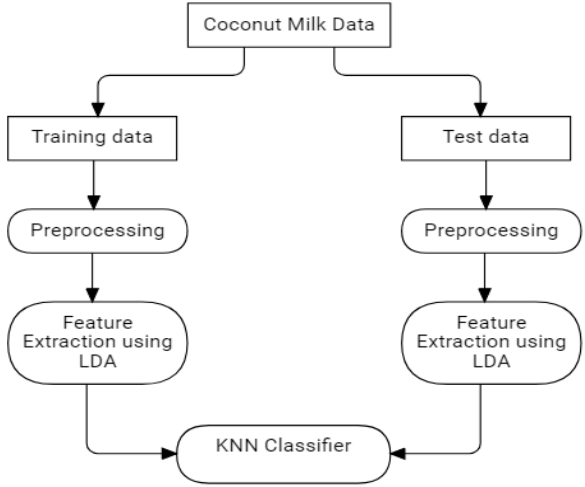

Fig. 2. Block diagram of the proposed system

### A. Preprocessing

The dataset used in this research contains infrared spectral data of two types of coconut milk. Previous work treated these two milk types separately and built prediction models for each of them. We applied the paired t-test to all authentic instances of the two milk types. The paired t-test is a statistical measure used to compare between means of observations from two samples. We compute it by calculating the mean and standard deviation of differences between the two samples. Then, we compute the p-value from the t distribution. If the p-value is equal to or less than 0.05, then the two samples are significantly different. The findings of the statistical test showed no significant difference between the two milk types. We treated the traditional and instant milk types as one type.

The next step of preparing the dataset was eliminating irrelevant spectral bands using backward feature elimination (BFE). In BFE, we evaluate the performance of the ML models using all features. Then, we remove one attribute and assess the ML classifiers' performance on the remaining variables. We repeat the process for all the features. As a result, we exclude the attributes that got the lowest accuracy. Experimentally, we found the spectral bands before the wavelength of 3150 nm and after 3840 nm were irrelevant. The models got the highest classification accuracy using the spectral bands in the interval between these two bands.

### B. Feature Extraction

In many classification tasks, feature extraction is critical because it enhances the performance of classification models by extracting only significant features. There are two approaches for extracting features: supervised and unsupervised. Supervised methods, such as LDA, require class labels to produce the attributes. The unsupervised algorithms, like Principal Component Analysis (PCA), do not demand class labels. We used the LDA technique to extract features from the coconut milk spectral dataset in this study. We used LDA because it had previously produced positive findings in a hyperspectral imaging dataset [7]. The LDA technique translates features into a lower-dimensional space that increases the ratio of inter-class variance to intra-class variation, ensuring high class separation [8]. There are two methods for implementing LDA: class dependent and class independent LDA. The goal of class-dependent LDA is to optimize the between-class to the within-class ratio for all classes. The class independent LDA aims to maximize the total between-class variance to within-class variation ratio. Because it just involves the solution of a generalized eigenvalue system, LDA has the advantage of being rapid [9]. It also works for binary-class and multi-class problems, and it may extend to nonlinear LDA by adding a quadratic kernel. The new LDA features are calculated from the original features using the following steps:

Step 1: Compute the mean value $\mu$ of all instances $x_i, i = 1 \dots N$ in the dataset using equation 1.

$$\mu = \frac{1}{N} \sum_{i=1}^{N} x_i \qquad (1)$$

Step 2: Compute the mean value of each class in the dataset using equation 2.

$$\mu_j = \frac{1}{n_j} \sum_{x_i \in \omega_j} x_i \qquad (2)$$

Step 3: Compute the within-class and between-class scatter matrices $S_W$, $S_B$ using equations 3 and 4 where $p_j$ is the prior probability of $jth$ class.

$$S_w = \sum_j p_j \times (x_j - \mu_j)(x_j - \mu_j)^T \qquad (3)$$

$$S_B = \sum_j (\mu_j - \mu) \times (\mu_j - \mu)^T \qquad (4)$$

Step 4: Compute the transformation matrix using equation 5.

$$W = S_W^{-1} S_B \qquad (5)$$

Step 5: Calculate the eigenvalues $\lambda$ and eigenvectors of $W$ and construct a matrix $T$ containing the eigenvectors that correspond to the largest eigenvalues.

Step 6: Transform the original features matrix $X$ into the new features matrix $Y$ using equation 6.

$$Y = X \times T \qquad (6)$$

In this study, we compare the performance of the LDA features to the efficiency of features reduced by an unsupervised algorithm, such as PCA.

## C. Classification

The last phase of the proposed system is classification. In this stage, we employed KNN to classify the coconut milk samples into pure, adulterated10, or adulterated20, based on their features. The KNN model is a supervised ML classification algorithm. It is memory-based and does not require the fitting of a model [10]. In this ML classification technique, we locate the k training points x(r), r = 1, …, k closest to a query point x0, then classify them using a majority vote among the k neighbors given x0. A wide range of classification problems, including food quality assessment, used KNN successfully. When the decision boundary is irregular, the KNN classifier is often successful. We used the KNN classifier because it performed well on spectroscopic data in previous works.

In the present research, we compare the performance of the KNN classifier to the efficiency of another popular ML algorithm, which is SVM. We chose the SVM classifier since it achieved excellent results in previous work on spectral data. The SVM model is an ML classifier that produces an optimal hyperplane in a transformed higher-dimensional feature space to separate classes with the fewest possible errors [11]. We used two SVM models, one with a linear kernel and another with a radial basis function (RBF) kernel function.

## D. Performance Evaluation

We evaluated the performance of the machine learning classifiers (KNN and SVM) using the stratified cross-validation method with five folds. In each fold, we divided the dataset into training and test sets. We trained the classifiers using the training set and evaluated their performance using the test set. We used balanced accuracy as the performance metric of the classifiers. The balanced accuracy is a robust performance metric, especially for imbalanced datasets [12]. We can compute the balanced accuracy as the average of the sensitivity and specificity, as shown in equation 7.

$$Balanced\ Accuracy = \frac{TP}{TP+FN} + \frac{TN}{TN+FP} \qquad (7)$$

We represent the number of correctly classified positive and negative examples by the true positive (TP) and true-negative (TN), respectively. We depict the number of wrongly classified positive and negative instances by the false positive (FP) and false-negative (FN).

## III. RESULTS

In the present study, we used two popular supervised machine learning (ML) classification algorithms, namely KNN and SVM, to classify coconut milk samples into three categories. They are non-adulterated, adulterated at 10% water, and adulterated at 20% water. Table I shows the performance of the ML algorithms using the original features, PCA features, and LDA features. The machine learning models performed poorly using the original and PCA features, as seen in the table. Using LDA features, the classifiers performed very well. The results reveal that the KNN classifier outperformed other classifiers. It achieved a classification accuracy of 93.33% for classifying coconut milk spectral instances into authentic, adulterated with 10% water, or contaminated with 20% water. Figure 3 visualizes the classifiers' performance using the original features, PCA features, and LDA features.

TABLE I. PERFORMANCE OF THE ML CLASSIFIERS FOR DIFFERENT FEATURE SETS

| ML Model | Original features | PCA features | LDA features |
|---|---|---|---|
| KNN | 44.44% | 26.67% | 93.33% |
| Linear SVM | 37.78% | 37.78% | 86.67% |
| RBF SVM | 32.22% | 40% | 91.11% |

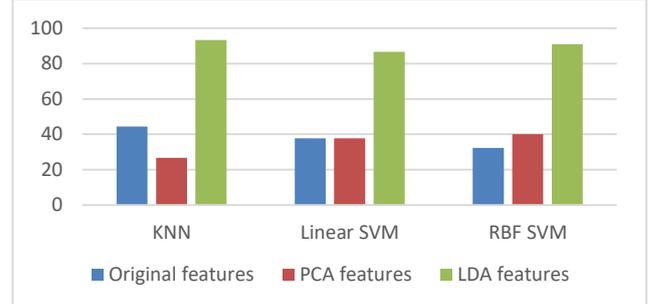

Fig. 3. Performance of the ML classifiers for different feature sets

Table II shows the classification accuracy of the classifiers' accuracy for each class in the dataset using LDA features. We observe that the classifiers' performance differs for each group, except for the authentic category, where all classifiers achieved the same classification accuracy of 86.67%. According to the results, the adulterated20 class was the most accurately classified.

TABLE II. PERFORMANCE OF THE CLASSIFIERS FOR EACH CLASS USING LDA FEATURES

| ML Model | Class | Authentic | Adulterated10 | Adulterated20 |
|---|---|---|---|
| KNN | 86.67% | 93.33% | 100% |
| Linear SVM | 86.67% | 86..67% | 86.67% |
| RBF SVM | 86.67% | 86.67% | 100% |

Figure 4 shows the classification accuracy of the KNN model for different values of the KNN parameter k. The figure shows KNN achieved the highest performance using five nearest neighbors.

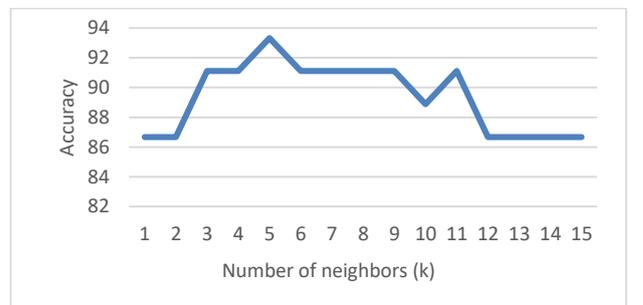

Fig. 4. Accuracy of KNN classifier for different values of the parameter k

Figure 5 displays the proposed system's graphical user interface. The IR spectral data of a coconut milk sample is entered into the system by a user. When the user presses the "Test Sample" button, the system removes the extraneous bands from the spectral signal. The algorithm then derives LDA features from the sample's spectral data and feeds them into a KNN classifier. Finally, the KNN model classifies the features into authentic, adulterated10, or adulterated20.

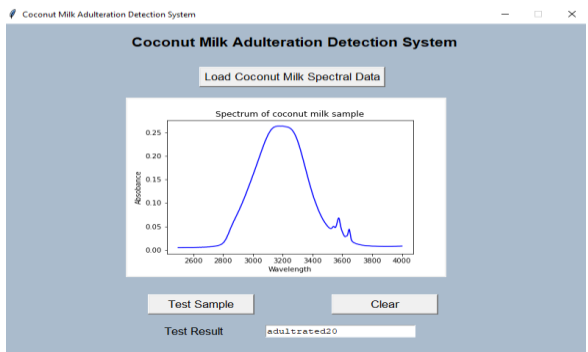

Fig. 5. The Graphical user interface of the proposed system

## IV. DISCUSSION

The classifiers underperformed using the original features since the dataset contained many irrelevant data. Likewise, the performance of the classifiers did not improve using the reduced features using PCA because it could not separate the classes well, as shown in figure 6.

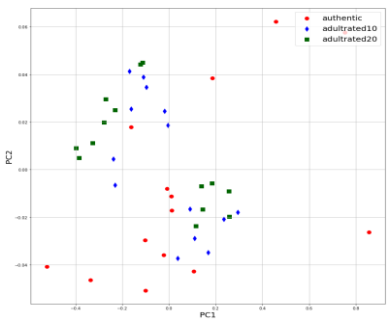

Fig. 6. Projection of the spectral data on the first two principal components

The classifiers performed tremendously using the LDA features since the LDA algorithm increased the class separation, as shown in figure 7.

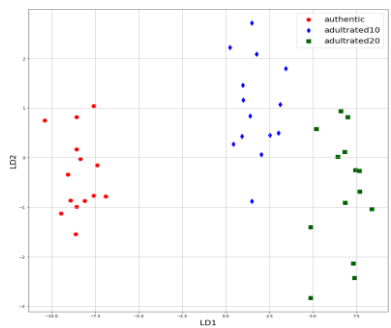

Fig. 7. Projection of the spectral data on the first two linear discriminant components

The SVM classifier with the RBF kernel outperformed the linear classifier because of its ability to model nonlinear relationships in the dataset. All the classifiers achieved the same accuracy for the authentic samples since the samples were apart from other specimens, as shown in figure 7. Using spectral data from other electromagnetic spectrum regions, such as visible near-infrared and ultraviolet, may improve the current detection system performance.

## I. CONCLUSION AND FUTURE WORK

In this research, we proposed a system for detecting coconut milk adulteration with water. Preprocessing, feature extraction, and classification were the main steps of the proposed adulteration detection system. In the preprocessing step, we removed irrelevant spectral bands from the dataset. The feature extraction step employed the LDA algorithm. We used the KNN model in the classification step. We trained the proposed system and evaluated its performance on a public dataset comprising FTIR spectra of various coconut milk samples. The proposed system successfully detected and quantified adulteration in coconut milk with a classification accuracy of 93.33%. Findings confirm the effectiveness of FTIR spectroscopy and ML in detecting adulteration in coconut milk. These techniques provide a fast, automatic, and nondestructive approach for coconut milk quality assessment. Our next step is applying the proposed method for detecting and quantifying adulteration in other food products.